\documentclass[sigconf]{acmart}

\copyrightyear{2024}
\acmYear{2024}
\setcopyright{rightsretained}
\acmConference[FAccT '24]{The 2024 ACM Conference on Fairness, Accountability, and Transparency}{June 3--6, 2024}{Rio de Janeiro, Brazil}
\acmBooktitle{The 2024 ACM Conference on Fairness, Accountability, and Transparency (FAccT '24), June 3--6, 2024, Rio de Janeiro, Brazil}
\acmDOI{10.1145/3630106.3658908}
\acmISBN{979-8-4007-0450-5/24/06}


\usepackage{amsthm}
\usepackage{xspace}
\usepackage{natbib}

\newcommand{\hY}{\hat{Y}}
\newcommand{\cD}{\mathcal{D}}
\newcommand{\cA}{\mathcal{A}}
\newcommand{\R}{\mathbb{R}}
\newcommand{\E}{\mathbb{E}}
\renewcommand{\Pr}{\mathbb{P}}
\newcommand{\iid}{i.i.d.\@\xspace}
\newcommand{\hmu}{\hat{\mu}}
\newcommand{\hsigma}{\hat{\sigma}}
\newcommand{\htau}{\hat{\tau}}
\newcommand{\mse}{\textsf{MSE}}
\newcommand{\bias}{\textsf{Bias}}
\newcommand{\var}{\textsf{Var}}
\newcommand{\one}{\boldsymbol{1}}
\newcommand{\zero}{\boldsymbol{0}}
\newcommand{\boot}{\text{boot}}
\newcommand{\cN}{\mathcal{N}}
\newcommand{\Bernoulli}{\textup{Bernoulli}}
\newcommand{\model}{\text{model}}
\newcommand{\figsqueeze}{\vspace{-6pt}}
\newcommand{\tabsqueeze}{\vspace{-3pt}}

\newcommand{\cE}{\mathcal{E}}
\newcommand{\vtheta}{\boldsymbol{\theta}}
\newcommand{\hvtheta}{\boldsymbol{\hat{\theta}}}
\newcommand{\vphia}{\boldsymbol{\phi}^a}
\newcommand{\phia}[1]{\phi^a_{#1}}
\newcommand{\htheta}{\hat{\theta}}
\newcommand{\inprod}{\mathbin{\cdot}}
\newcommand{\race}{\textit{race}\xspace}
\newcommand{\rc}{\textit{rc}}
\newcommand{\age}{\textit{age}\xspace}
\newcommand{\gender}{\textit{gender}\xspace}
\newcommand{\expl}{\textit{expl}\xspace}
\newcommand{\sens}{\textit{sens}\xspace}
\newcommand{\EB}{\textsc{eb}}
\newcommand{\JS}{\textsc{js}}

\newcommand{\given}{\mathrel{\vert}}
\newcommand{\bigGiven}{\mathrel{\bigm\vert}}
\newcommand{\BigGiven}{\mathrel{\Bigm\vert}}

\newcommand{\BiggGiven}{\mathrel{\Biggm\vert}}
\newcommand{\set}[1]{\{#1\}}

\newcommand{\card}[1]{\lvert#1\rvert}

\newcommand{\norm}[1]{\lVert#1\rVert}

\newcommand{\bigParens}[1]{\bigl(#1\bigr)}
\newcommand{\BigParens}[1]{\Bigl(#1\Bigr)}

\newcommand{\BiggParens}[1]{\Biggl(#1\Biggr)}
\newcommand{\Parens}[1]{\left(#1\right)}
\newcommand{\bigBracks}[1]{\bigl[#1\bigr]}
\newcommand{\BigBracks}[1]{\Bigl[#1\Bigr]}

\newcommand{\BiggBracks}[1]{\Biggl[#1\Biggr]}

\newcommand{\bracks}[1]{[#1]}

\newcommand{\sel}{\textup{SEL}}
\newcommand{\fnr}{\textup{FNR}}
\newcommand{\fpr}{\textup{FPR}}
\newcommand{\ppv}{\textup{PPV}}
\newcommand{\wer}{\textup{WER}}
\newcommand{\acc}{\textup{ACC}}
\newcommand{\auc}{\textup{AUC}}
\newcommand{\weru}{\textit{wer}}
\newcommand{\eps}{\varepsilon}

\newcommand{\Eq}[1]{Eq.~\eqref{eq:#1}}

\theoremstyle{definition}
\newtheorem{example}{Example}

\begin{document}

\title[]{A structured regression approach for evaluating model performance across intersectional subgroups}

\author{Christine Herlihy}
\authornote{Both authors contributed equally to this research.}
\email{cherlihy@umd.edu}
\affiliation{%
  \institution{University of Maryland}
  \city{College Park}
  \country{USA}
}
\author{Kimberly Truong}
\authornotemark[1]
\email{truonkim@oregonstate.edu}
\affiliation{%
  \institution{Oregon State University}
  \city{Corvallis}
  \country{USA}
}
\author{Alexandra Chouldechova}
\email{alexandrac@microsoft.com}
\affiliation{%
  \institution{Microsoft Research}
  \city{New York}
  \country{USA}
}

\author{Miroslav Dud\'ik}
\email{mdudik@microsoft.com}
\affiliation{%
  \institution{Microsoft Research}
  \city{New York}
  \country{USA}
}

\begin{abstract}
Disaggregated evaluation is a central task in AI fairness assessment, where the goal is to measure an AI system's performance across different subgroups defined by combinations of demographic or other sensitive attributes. The standard approach is to stratify the evaluation data across subgroups and
compute performance metrics
separately for each group. However, even for moderately-sized evaluation datasets, sample sizes quickly get small once considering intersectional subgroups, which greatly limits the extent to which intersectional groups are included in analysis.
In this work,
we introduce a structured regression approach to disaggregated evaluation that we demonstrate can yield reliable system performance estimates even for very small subgroups. We provide corresponding inference strategies for constructing confidence intervals and explore how goodness-of-fit testing can yield insight into the structure of fairness-related harms experienced by intersectional groups. We evaluate our approach on two publicly available datasets, and several variants of semi-synthetic data. The results show that our method is considerably more accurate than the standard approach, especially for small subgroups, and demonstrate how goodness-of-fit testing helps identify the key factors that drive differences in performance.\looseness=-1
\end{abstract}

\maketitle

\section{Introduction}
\label{sec:intro}

A core task when assessing the fairness of an AI system is measuring its performance across different subgroups defined by combinations of demographic or other sensitive attributes.  Many of the best-known studies of algorithmic bias are grounded in this type of analysis~\citep{buolamwini2018gender,obermeyer2019dissecting,koenecke2020racial,propublica2016,sweeney2013discrimination},
including \citeauthor{buolamwini2018gender}'s
Gender Shades
study~\citep{buolamwini2018gender}, which found that commercial gender classifiers have much higher error rates for darker-skinned women than other groups, and  \citeauthor{obermeyer2019dissecting}'s study \citep{obermeyer2019dissecting} finding bias in commercial algorithms used to guide healthcare decisions.\looseness=-1

In their work formalizing this type of analysis,
\citet{barocas2021designing} introduce the term \textit{disaggregated evaluation} to refer to this task.  The authors draw attention to the many decisions that implicitly shape any given disaggregated evaluation: from who will be involved, to what data will be used, which statistical approach taken, and what kinds of inferences drawn.  In our work, we focus on the question of statistical methodology given an available dataset and pre-determined subgroups and performance metrics of interest.  We introduce a method for estimating performance across subgroups that we show (i) is more accurate than approaches taken in standard practice; and (ii) can provide greater insight
into which factors drive observed variation in performance. We do so through careful adaptations of well-established techniques rather than development of entirely novel statistical methods.

The ``standard approach'' to disaggregated evaluation proceeds by stratifying the evaluation data across subgroups and then conducting inference (i.e., computing performance metrics, confidence intervals, or other statistics) separately for each group.  The primary challenge when applying this approach comes from small sample sizes.  Even for moderately-sized evaluation datasets, sample sizes quickly get small once considering intersectional subgroups.  For instance, in a medical diabetes mellitus dataset we use later in the paper, we have a 5000-patient evaluation dataset, of which 2689 patients are female, 620 are female \textit{and} over age 80, but only 6 are female, over age 80, \textit{and} Hispanic. Indeed, of the 32 distinct gender-age-race/ethnicity subgroups that can be formed in the data, 8 (i.e., 25\%) have fewer than 10 observations, and nearly half have fewer than 25 observations. Inference based on so few observations is often uninformative, and may be unreliable. In practice, subgroups that are too small tend to be either excluded from analysis or merged with other small but potentially heterogeneous subgroups to form higher-level ``catch-all'' categories (e.g., ``other'').
These practices greatly limit the extent to which intersectional groups are even considered in many disaggregated evaluations.
As a consequence, standard assessments may fail to surface fairness-related harms that could disproportionately affect intersectional subgroups~\cite{crenshaw1989demarginalizing}, which in turn means that steps to mitigate those harms might not be taken.\looseness=-1

In this work, we introduce a structured regression approach to disaggregated evaluation that we demonstrate can yield reliable system performance estimates even for very small subgroups (e.g., for groups with fewer than 25 observations).  We also provide corresponding inference strategies for constructing confidence intervals for the subgroup-level performance estimates.  We then demonstrate how goodness-of-fit testing can provide insight into the structure of fairness-related harms experienced by intersectional groups
and also identify situations where observed variation in performance is attributable to benign factors.  Lastly, we present results on two publicly available datasets, and several variants of semi-synthetic data.  The results show that our method is considerably more accurate than the standard approach, especially for small subgroups.  They further show that our method outperforms
more statistically sophisticated
baselines, including the \emph{model-based metrics} method introduced by \citet{miller2021modelbased}, while also offering additional advantages.  We conclude by discussing limitations and future directions.

\section{Background and Related Work}
\label{sec:background}

In their taxonomy of sociotechnical harms of algorithmic systems, \citet{shelby2023sociotechnical} identify five high-level categories of harm: representational, allocative, quality of service, interpersonal, and social system. Our work contributes to the broader literature characterizing and assessing \textit{allocative} and \textit{quality-of-service} harms that can result from the use of algorithmic systems.  Allocative harms, first discussed by \citet{barocas2017problem}, occur when systems produce an inequitable distribution of information, opportunity, or resources across groups. As a running example,
we consider a hypothetical setting in which a model trained to predict 30-day hospital readmission is used to prioritize high-risk patients for more intensive post-discharge care.  Allocative harms might occur if certain subgroups of patients are disproportionately under-prioritized for more intensive care (i.e., have low selection rates) or are under-selected relative to their observed rate of readmission (i.e., have high false negative rates).

Quality-of-service harms occur when algorithmic systems underperform for certain socially salient groups of users~\cite{weerts2023fairlearn,shelby2023sociotechnical}.  We examine quality-of-service harms across race and gender groups in the context of commercial automated speech recognition (ASR) systems using data previously analyzed by \citet{koenecke2020racial}.  Specifically, we assess whether there is significant variation in the word error rate (WER) of the ASR systems across intersectional race and gender subgroups.\looseness-1

The term ``intersectionality'' was introduced by \citet{crenshaw1989demarginalizing} to describe the distinct patterns of discrimination and disadvantage experienced by Black women, which she argued cannot be understood in terms of race or gender discrimination alone.  In recent years, algorithmic fairness research has examined intersectional bias from many perspectives.  This includes work introducing quantitative metrics intended to capture notions of intersectional fairness, such as subgroup fairness \cite{kearns2018preventing}, differential fairness \cite{foulds2020intersectional, foulds2020bayesian}, and multi-calibration~\cite{hebert2018multicalibration}, along with learning algorithms for estimating and achieving these criteria.  \citet{wang2022towards} study ``predictivity differences'' across intersectional subgroups, and discuss limitations of existing summary statistics (such as the maximum disparity across all groups) in capturing meaningful notions of intersectional harm.  Our work differs from this literature because we are specifically interested in the task of disaggregated evaluation. This entails estimating and reporting system performance for each intersectional subgroup, rather than computing a particular fairness metric or learning a fairness-constrained model.

Our work most directly contributes to the growing literature introducing more sample-efficient methods for conducting disaggregated evaluations.  This literature includes methods that leverage unlabelled data in model evaluation \cite{ji2020can, ji2020active, chouldechova2022unsupervised}; methods that bound or approximate performance for intersectional subgroups using marginal statistics \cite{molina2022bounding}; and synthetic data augmentation approaches \cite{van2023can}.  In work more closely related to the spirit of our structured regression approach, \citet{piralta2021active} introduce the \emph{attributed accuracy assay} (AAA) method, which models the accuracy of a model as a function of sensitive attributes and other features via a Gaussian process (GP).  While we do not rely on GPs, we do proceed similarly by modeling the accuracy (or error) of a given model. Whereas we are specifically concerned with fairness and disaggregated evaluation, \citet{piralta2021active} aim to produce an ``accuracy surface'' model that clients can use to estimate the performance of an existing model on their data.\looseness=-1

The most closely related work in recent literature is that of \citet{miller2021modelbased}, who introduce a Bayesian structured regression approach that they call \textit{model-based metrics} (MBM). Their method applies to AI models that produce a score (say to predict a risk of hospital readmission). By modeling the distribution of scores given select features and the observed outcome, they are able to make inference on any performance metric of interest, but the approach is not directly applicable to the evaluation of models that do not produce classification scores (e.g., MBM does not directly apply to the evaluation of WER in ASR systems). Unlike the MBM approach, we model the target metric
directly and fit separate models for each performance metric of interest. Our experiments show that our method yields more accurate estimates than MBM (see \textsection\ref{sec:exp:diabetes}).\looseness=-1

Our approach is also related to the classical line of research on normal means estimation, originating with the James-Stein (JS) estimator~\cite{james1961estimation,stein1956inadmissibility}. The JS estimator works by shrinking standard estimates towards zero (or some other constant). This leads to a substantial decrease in variance, which outweighs a moderate increase in bias, and yields a more accurate estimator.
The empirical Bayes (EB) approach~\citep{efron1973stein} also leads to a form of shrinkage, but its motivation is different. It posits a hierarchical Bayesian model and estimates metric values by posterior means, while fixing prior hyperparameters to their point estimates.
Our estimator works by optimizing bias--variance trade-off similar to JS, but it enjoys additional advantages compared with JS and EB: availability of confidence intervals  and flexibility to incorporate covariate information.
In our experiments we show that our approach can outperform  JS and EB in terms of estimation accuracy
(see \textsection\ref{sec:exp:diabetes}).\looseness=-1

\section{Problem setting}
\label{sec:setting}

Our goal is to assess fairness-related harms of an AI system by evaluating its performance on intersectional subgroups of users specified by $k\ge 2$ sensitive attributes (like race and gender), taking values in finite sets $\cA_1,\dotsc,\cA_k$. The set of all possible $k$-tuples of sensitive-attribute values is denoted $\cA=\cA_1\times\dotsb\times\cA_k$.

We assume that we have access to an evaluation dataset $S$, consisting of individuals described by tuples of the form $\smash{(X,A,Y,\hY)}$ sampled \iid from some underlying distribution $\cD$, where $X$ contains application-relevant information about the individual (e.g., the health history of a patient), $A\in\cA$ is a $k$-tuple of sensitive attributes, $Y$ is an observed outcome variable (e.g., whether the patient was readmitted within 30 days of discharge), and $\hY$ is an output produced by the AI system (e.g., a score used for prioritizing patients into post-discharge care).

For any $k$-tuple $a\in\cA$, we write $a[1],\dotsc,a[k]$ to denote its components. In the ASR example below, we consider two sensitive attributes, \race and \gender, with domains $\cA_1=\set{\text{Black}, \text{white}}$ and $\cA_2=\set{\text{male}, \text{female}}$.
For $a=(\text{Black},\text{female})$, we then have ${a[1]=\text{Black}}$ and $a[2]=\text{female}$.
When possible, we use mnemonic indices for components of $a$ and write $a[\race]$ and $a[\gender]$ to mean $a[1]$ and $a[2]$, and similarly $\cA_\race$ and $\cA_\gender$ to mean $\cA_1$ and $\cA_2$.\looseness=-1

For each $a\in\cA$, we define $\cD_a$ to be the distribution of individuals with $A=a$, so $\cD_a$ is the conditional distribution $\cD(X,A,Y,\hY\given A=a)$, representing an intersectional group. Let $\Delta$ denote the set of all probability distributions over tuples $(X,A,Y,\hY)$, so $\cD\in\Delta$ and also $\cD_a\in\Delta$ for all $a\in\cA$. A \emph{performance metric} is a function $m:\Delta\to\R$ that maps a probability distribution over tuples $(X,A,Y,\hY)$ into a real number.
For example, if the underlying AI system performs binary classification, so $Y,\hY\in\set{0,1}$, we could measure its performance using \emph{accuracy}, defined, for any $p\in\Delta$, as
\[
  \acc(p)=\Pr_{p}\bracks{Y=\hY},
\]
where $\Pr_p[\cdot]$ is the probability of an event with respect to $p$. The overall system performance is then quantified by $\acc(\cD)$ and the performance on a group $a\in\cA$ by $\acc(\cD_a)$.

Given a performance metric $m$, the goal of \emph{disaggregated evaluation} is to estimate,
for all $a\in\cA$, the values
\[
  \mu_a = m(\cD_a).
\]

Our only source of information about $\cD$ is the evaluation dataset $S$ of size $n=\card{S}$, sampled \iid from $\cD$. The \emph{standard approach} to disaggregated evaluation splits the dataset $S$ into groups $S_a = \set{(X,A,Y,\hY)\in S:\:A=a}$ of size $n_a=\card{S_a}$,
and then evaluates $m$ on each $S_a$ (or, more precisely, on the probability distribution that puts an equal probability mass on each data point in $S_a$).
We denote the resulting \emph{standard estimates} by
\begin{equation*}
  Z_a = m(S_a).
\end{equation*}
For example, if $m$ is accuracy, then
\[
  \smash[t]{Z_a = \acc(S_a) = \frac{1}{n_a}\sum_{(X,A,Y,\hY)\in S_a}\one\set{Y=\hY},}
\]
where $\one\set{\cdot}$ is an indicator equal to 1 if its argument is true and 0 if it is false.

We next connect this abstract framework to two concrete scenarios already mentioned in \textsection\ref{sec:background}.

\begin{example}[Diabetes]
\label{ex:diabetes}
We consider an AI system that refers high-risk patients into a post-discharge care program. We wish to assess allocative harms of this system. To explore this scenario,
we use a publicly available dataset of diabetes patients developed by \citet{strack2014impact}. The dataset contains information about patient hospital visits, including whether each patient was readmitted within 30 days after discharge. We use the readmission as a proxy for whether the patient should be recommended for the care program.

Each data point corresponds to a patient admission, where $X$ describes the patient history and hospital tests; $A$ describes the patient's \race, \gender, and (binned) \age; $Y\in\set{0,1}$ indicates whether the patient was readmitted within 30 days after discharge; and $\hY\in[0,1]$ is the score produced by the AI system that has been trained to predict $Y$.
We assume that the hospital uses a threshold $r$, and patients with $\hY\ge r$ are automatically referred into the care program.\looseness=-1

One type of allocative harm occurs when a subgroup of patients is disproportionately under-prioritized, i.e., if a subgroup has a low \emph{selection rate}, denoted as
\[
  \sel(\cD_a) =
  \Pr_{\cD_a}\bracks{\hY\ge r}.
\]

We also consider a second type of harm, which occurs when a subgroup of patients experiences a disproportionately large rate of false negatives
(i.e., many of those patients that should be recommended are not),
measured by the \emph{false negative rate}
\[
  \fnr(\cD_a) =
  \Pr_{\cD_a}\bracks{\hY<r\given Y=1}.
\]
\end{example}

\begin{example}[ASR]
\label{ex:ASR}
To assess quality-of-service harms of an ASR system, we use a dataset from \citet{koenecke2020racial}, consisting of audio snippets (of length between 5s and 50s) spoken by various speakers. In the dataset, $X$ describes properties of the snippet (like duration in seconds),
$A$ has two components corresponding to the speaker's \race and \gender, $Y$ is the ground-truth transcription of the snippet, and $\hY$ is the transcription provided by the AI system.

The quality-of-service harms occur when the system underperforms for a subgroup of users. The performance is evaluated by the \emph{word error rate}
\[
  \wer(\cD_a) = \E_{\cD_a}[\weru(\hY,Y)],
\]
where $\weru$ is a snippet-level word error rate defined as
\[
  \weru(\hY,Y) = \frac{\textit{subst}+\textit{del}+\textit{ins}}{\card{Y}},
\]
where $\textit{subst}$, $\textit{del}$, and $\textit{ins}$ is the number of word substitutions, deletions, and insertions in $\hY$ compared with the ground truth $Y$, and $\card{Y}$ is the number of words in $Y$.
\end{example}

\begin{figure*}%
\centering%
\includegraphics{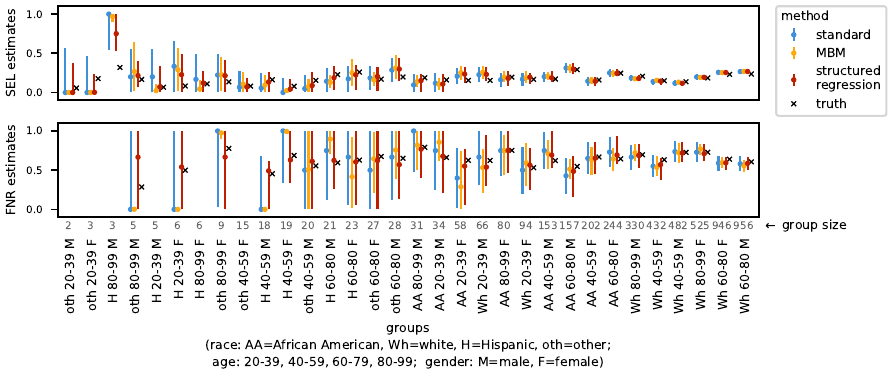}

\figsqueeze
\caption{\emph{Point estimates and 95\% confidence intervals of selection rate (SEL) and false negative rate (FNR) on diabetes data.} Confidence intervals of the standard estimator are calculated using pooled variance (see Eqs.~\ref{eq:pooled} and~\ref{eq:pooled:CI}).}
\label{fig:errors-sel-fnr}
\end{figure*}

To quantify the accuracy of an estimator, like the standard estimator introduced above, we often use \emph{mean squared error} (MSE). We use a modified definition of MSE that accounts for the fact that estimates like $Z_a=m(S_a)$ are sometimes undefined, for
instance,
when the metric $m$ is defined as a conditional probability, like $\fnr$ in Example~\ref{ex:diabetes}, and the set $S_a$ has no samples that satisfy the condition (e.g., no samples with $Y=1$ in case of $\fnr$).
For an estimator $\hmu$ of a quantity~$\mu$, let $\cE$ denote the event that $\hmu$ is defined. The \emph{bias}, \emph{variance}, and \emph{mean squared error} (MSE) of $\hmu$ are defined as
\begin{align*}
  \bias(\hmu)&=
  \E[\hmu\given\cE]-\mu,
\\
  \var(\hmu)&=
  \E\bigBracks{\bigParens{\hmu-\E[\hmu\given\cE]}^2\bigGiven\cE},
\\
  \mse(\hmu)&=
  \E\bigBracks{\bigParens{\hmu-\mu}^2\bigGiven\cE},
\end{align*}
where the expectations are with respect to the data-generating process giving rise to the dataset used to calculate $\hmu$ (which is itself a random variable).
An estimator with bias equal to zero is called \emph{unbiased}.

Mean squared error decomposes into bias and variance terms as
\begin{equation*}
  \mse(\hmu)
  =
  \bigBracks{\bias(\hmu)}^2 +\var(\hmu),
\end{equation*}
so for unbiased estimators, mean squared error is equal to variance.

Throughout the paper, we assume that the standard estimates $Z_a$ are unbiased. Writing this condition in terms of the metric~$m$,
we assume that for all $\cD\in\Delta$ and all $n\ge 1$, the performance metric $m$ satisfies
\begin{equation}
\label{eq:assume:unbiased}
   \E_{S\sim\cD^n}[m(S)\given\text{$m(S)$ is defined}] = m(\cD),
\end{equation}
which is true for all the metrics in this paper.
Substituting $\cD_a$ for $\cD$ and $n_a$ for $n$ in \Eq{assume:unbiased} implies that $\E[Z_a\given\text{$Z_a$ is defined}]=\mu_a$. In the rest of the paper we drop conditioning on the events like ``$Z_a$ is defined,'' and just write $\E[Z_a]=\mu_a$ for simplicity.

Since the standard estimates $Z_a$ are unbiased, their MSE is equal to their variance, which typically scales as $O(1/n_a)$, the inverse of the number of samples in the group. Thus, standard estimates are accurate when $n_a$ is large, but less so when $n_a$ is small. Unfortunately, even for moderately sized evaluation datasets, the sizes of intersectional groups can be quite small. In Figure~\ref{fig:errors-sel-fnr}, we show standard estimates of $\sel$ and $\fnr$ on diabetes data (alongside estimates produced by methods introduced later in the paper), along with group sample sizes, $n_a$ (almost half of which are less than 25).

\section{Structured regression approach}

We next develop a structured regression (SR) approach, which seeks to overcome the main shortcoming of the standard estimator:
its large variance for small groups. Our approach builds on two main ideas. First, we \textit{enable variance reduction} by leveraging information across \emph{all data points}, not just data points in $S_a$, to estimate $\mu_a$. We do this by pooling the data across related groups, for example, across intersectional groups that agree in one of their attributes (like \age), and by using additional explanatory variables (like $X$), both of which is accomplished by fitting a regression model for $\mu_a$'s, viewing $Z_a$'s as observations thereof. Second, we ensure our regression model is always correctly specified (and thus can produce an unbiased estimate) by including all $k$-way interaction features.  This means our regression model can recover the standard unbiased estimates $Z_a$ as a special case. Regularization is used to optimize the bias--variance trade-off between the high-variance standard estimator and a high-bias (but low-variance) constant estimator.\looseness=-1

To start, since the standard estimates are unbiased, i.e., $\E[Z_a]=\mu_a$, we can write
\[
  Z_a = \mu_a + \eps_a
\]
for all $a\in\cA$, where $\eps_a$'s are independent random variables with $\E[\eps_a]=0$. We denote the variance of $Z_a$ as $\sigma^2_a=\var(Z_a)=\E[\eps_a^2]$. In order to estimate $\mu_a$, we consider a linear model of the form
\[
  \mu_a = \theta_0 + \vtheta\inprod\vphia
\]
for all $a\in\cA$, where $\vphia\in\R^d$ is the feature vector describing the group $a$, and $\theta_0\in\R$, $\vtheta\in\R^d$ are the parameters of the linear model. It remains to specify how to define $\vphia$, how to fit the parameters $\theta_0$ and $\vtheta$, and how to estimate $\sigma_a$.

\begin{figure*}
\centering%
\includegraphics{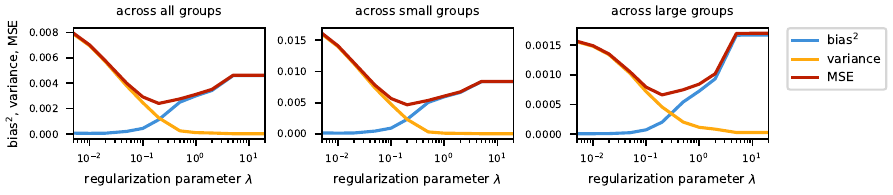}%

\figsqueeze
\caption{\emph{Bias--variance trade-off of structured regression estimates of selection rate (SEL) on diabetes data.} Averaged across all groups, small groups (size at most 25), and large groups (size above 25), across 100 draws of evaluation dataset. The scale of the MSE is different for different group sizes, but the minimum MSE is attained around the same value of $\lambda$, thanks to the weighting of the training loss.\looseness=-1}
\label{fig:bias-var-sel}
\end{figure*}

\subsubsection*{Defining feature vectors $\vphia$.}

The coordinates of $\vphia$ are referred to as features and denoted as $\phia{j}$ for $j$ from some suitable index set. We allow features to be linearly dependent.
We consider the following types of features:

  \begin{enumerate}
  \item\emph{Sensitive features.} These are derived directly from $a$. We always include group-identity indicators for all the groups $a'\in\cA$, yielding features of the form $\phia{a'}=\one\set{a=a'}$. This allows the linear model to express any combination of values~$\mu_a$. Additionally, in order to pool information across related groups, we also define indicators for individual attribute values, that is, features of the form $\phia{i,v}=\one\set{a[i]=v}$ for $i\in\set{1,\dotsc,k}$ and $v\in\cA_i$. In our diabetes example, there are three sensitive attributes: \race, \age, and \gender, with $\card{\cA_\race}=4$, $\card{\cA_\age}=4$, and $\card{\cA_\gender}=2$, so $\card{\cA}=4\cdot4\cdot2=32$. We use a total of 42 sensitive features: 32 group-identity indicators, 4~indicators of \race, 4~indicators of \age, and 2~indicators of \gender. An example of a group-identity indicator is $\phia{\text{(Hispanic,80--99,female)}}$ and an example of a sensitive-attribute indicator is $\phia{\race,\text{Hispanic}}$.
  \item\emph{Explanatory features.} These are derived from $X$, $Y$, and possibly $\hat{Y}$. We first featurize $X$ using some real-valued functions $f_j(X)$, $j=1,\dots,\ell$, and then define explanatory features $\phia{j}=\E_{X\sim S_a}[f_j(X)]$ (i.e., the average of the feature for group $a$). Additionally, when $Y$ is categorical, we define features $\phia{y}=\Pr_{Y\sim S_a}[Y=y]$ measuring rates of different outcomes in the group $a$. In our diabetes example, we use 7 explanatory features: 5 are derived from individual-level features $f_j$, including, for example, the number of inpatient days of a given patient in the prior year; and 2 are of the form $\phia{y}$ corresponding to 2 possible values of~$Y$.
  \item\emph{Interaction terms.} It is also possible to consider various interaction terms, both among features of the same type (e.g., between \gender and \age indicators),
      or of different types (e.g., between the outcome $Y$ and \age).
  \end{enumerate}

\subsubsection*{Fitting the linear model.}
  We fit $(\theta_0,\vtheta)$ by lasso regression~\citep{tibshirani1996regression}, minimizing an $\ell_1$-penalized square loss. To improve the statistical efficiency of the estimator, loss for each group $a$ is weighted inversely proportional to the variance of $Z_a$. Intuitively, since our model can express the true $\mu_a$, we expect the square loss on each group to be on the order of the variance of $Z_a$, so inverse weighting ``equalizes the scale'' of losses across groups. The penalized loss is then
\begin{equation}
\label{eq:lasso:obj}
  L_\lambda(\theta_0,\vtheta)
  =
  \sum_{a\in\cA} \frac{1}{\sigma_a^2} \BigParens{\theta_0 + \vtheta\inprod\vphia - Z_a}^2 + \lambda\norm{\vtheta}_1,
\end{equation}
where $\lambda$ is the regularization hyperparameter. Denoting the minimizer of $L_\lambda$ (for a given $\lambda$) as $(\htheta_0,\hvtheta)$, we obtain
the \emph{structured regression estimates }$\hmu_a = \htheta_0 + \hvtheta\inprod\vphia$.

Tuning $\lambda$ allows us to navigate the bias--variance trade-off.
Because sensitive features include indicators of all values $a\in\cA$, when $\lambda=0$, the loss is minimized by $\hmu_a=Z_a$ (i.e., we recover the standard estimates). As $\lambda\to\infty$, the optimization returns $\hvtheta\approx\zero$. Fixing $\hvtheta=\zero$ and optimizing only over the intercept term yields the constant solution
\[
  \hmu_a = \hmu_0\text{ for all }a \in \cA ,\quad
  \text{with}
  \quad
  \hmu_0 = \frac{\sum_{a\in\cA} Z_a/\sigma_a^2}{\sum_{a\in\cA} 1/\sigma_a^2},
\]
corresponding to a weighted average of $Z_a$'s. This solution has a small variance, but it may suffer from a large bias when the true values $\mu_a$ are far from identical. By tuning $\lambda$, we thus move between the high-variance standard estimate and the high-bias (but low-variance) constant estimate.
The mean squared error is typically minimized at some intermediate value of $\lambda$ (see Figure~\ref{fig:bias-var-sel}). We tune $\lambda$ by 10-fold cross-validation, where the individual folds are obtained by stratified sampling of the dataset $S$ with respect to the sensitive attribute tuple $A$.  We can expect our method to be \textit{consistent} (i.e., converge to the true values $\mu_a$ as all subgroup sizes $n_a$ grow) because the standard approach is consistent, and is included as the special case of our method with $\lambda = 0$.

\subsubsection*{Estimating variance $\sigma_a^2$.}

Variances $\sigma_a^2$ are needed to calculate weights in our optimization procedure. A simple approach is to estimate $\sigma_a^2$ separately on each dataset $S_a$
by using standard variance estimators (when available), or, more generically, by bootstrap. Unfortunately, for small sample sizes, these variance estimates themselves might be inaccurate.

To overcome this limitation, we posit a parametric model for variance, namely, $\sigma_a^2=\sigma^2/n_a$,
for some parameter~$\sigma$.
To estimate $\sigma$, we proceed in two stages. We first use bootstrap on each set $S_a$ to obtain the initial estimate of $\sigma_a^2$, which we denote $(\hsigma_a^\boot)^2$. Thus, $n_a(\hsigma_a^\boot)^2$ is the initial estimate of $\sigma^2$. We expect the variance of this estimate to be on the order $O(1/n_a)$. Taking a weighted average across groups, with weighting inversely proportional to $(1/n_a)$, yields our final estimator of $\sigma^2$, which translates into an estimator of $\sigma_a^2$:
\begin{align}
\notag
  \hsigma^2 &= \frac{\sum_{a\in\cA} n_a\cdot\bigBracks{n_a(\hsigma_a^\boot)^2}}{\sum_{a\in\cA} n_a},
\\
\label{eq:pooled}
  \hsigma_a^2&=\hsigma^2/n_a
\text{ for all $a\in\cA$.}
\end{align}
We refer to these as the \emph{pooled estimates} of variance. In our preliminary experiments, these performed better than the initial estimates $(\hsigma_a^\boot)^2$, particularly on small datasets.  We note that even if we severely misestimate the variances $\sigma^2_a$, our estimation method remains consistent. This is because estimates $\hsigma_a$ appear only as weights in the objective \eqref{eq:lasso:obj}.  Misspecifying the weights will negatively impact the efficiency of the estimator, but not its consistency.

\subsection{Confidence intervals}
\label{sec:CI}

So far we have focused on obtaining point estimates $\hmu_a$. However, in order for these estimates to be useful in practice, we also need to quantify our uncertainty about their values. We do so by using confidence intervals. For unbiased estimators, like the standard estimator $Z_a$, confidence intervals can be derived by estimating the variance and then using normal approximation, which works quite well for $Z_a$ with the pooled estimates of variance (see Appendix~\ref{appsec:pooled}).

This approach
does not work with lasso estimates, because they are biased---in fact, they achieve their improved accuracy \emph{by being biased}---and so simple confidence intervals derived from variance estimates or bootstrap percentiles
are too narrow. Fortunately, there is a rich literature on lasso-based confidence intervals~\citep{zhang2014confidence,van2014on,javanmard2014confidence}. We use the \emph{residual bootstrap lasso+partial ridge} (rBLPR) approach of~\citet{liu2020bootstrap}. As the name suggests, it is based on a two-stage \emph{lasso+partial ridge} (LPR) point estimator, which first runs lasso as a feature-selection method, and then fits a ridge regression model, which only penalizes the features that were not selected by lasso. The rBLPR method calculates confidence intervals for the LPR estimate by residual bootstrap (see \citep{liu2020bootstrap} for details).\looseness=-1

\subsection{Understanding structure of performance variation through goodness-of-fit testing}
\label{sec:gof-testing}

When presenting the results of disaggregated evaluations, the most common approach is to display point estimates and (sometimes) confidence intervals for every subgroup, as we see, for example, in Figure~\ref{fig:errors-sel-fnr}. While this type of a plot can be helpful in identifying groups that may experience poor performance or allocation, it does not provide a narrative for understanding how these harms accrue.  Goodness-of-fit testing can complement disaggregated evaluations by allowing us to answer questions such as:

\begin{enumerate}
    \item \emph{Do intersectional groups experience additive, sub-additive, or super-additive fairness-related harms?}  For example, when a model is found to perform poorly for Black women, is this explained by the model performing poorly for Black people and women, or are there additional sources of error specific to the intersectional group of Black women? An answer to this question can, for example, inform future collection of training data.

    \item \emph{Are there benign factors
        that explain a significant amount of the observed performance variation across groups?} For example, are observed differences in the performance of an ASR system attributable to systematically worse audio quality in the recordings for speakers from certain groups? Presence of such benign factors \emph{does not lessen the harm}, but the knowledge of the factors that drive performance differences can be used to design mitigations (for example, denoising algorithms targeted at specific types of sensors or noise characteristics).
\end{enumerate}

These types of questions can be framed as goodness-of-fit tests. We consider goodness-of-fit tests that compare two linear models: $M_0$, with fewer features, and $M_1$, with some additional features. Such a test asks whether the additional features included in model $M_1$ improve the goodness of fit compared with model $M_0$, where the goodness of fit is measured using the square loss as in \Eq{lasso:obj}. To answer the first question above, we can compare a model $M_0$, which includes only indicators of \race and \gender, with a model $M_1$, which also includes interaction terms. To answer the second question, we can compare a model $M'_0$, which only includes benign factors, with a model $M'_1$, which additionally includes indicators of \race, \gender, and \age.

While there are goodness-of-fit tests that have been designed for lasso regression~\citep{jankova2020goodness}, in this paper, we use standard $F$-tests designed for unregularized linear regression. In contrast to the foregoing discussion, we do not include features corresponding to the indicators of $a$ (because these would trivially yield the standard estimates $Z_a$ with $0$ residual sum-of-squares (RSS), which in this case corresponds to overfitting).

\section{Experiments}
\label{sec:experiments}

In this section, we evaluate the accuracy of point estimates and calibration of confidence intervals produced by our structured regression (SR) approach. We also demonstrate how goodness-of-fit tests can be used to provide insights about what drives the variation of performance across groups.

We compare SR with several baselines. First, there is the \textbf{standard estimator} $Z_a = m(S_a)$. We construct confidence intervals for $Z_a$ using normal approximation with pooled variance estimates $(\hsigma_a)^2$ from \Eq{pooled}. Given a confidence level $\gamma$ (say 95\%), or a significance level $\alpha=1-\gamma$ (say 5\%), we use the confidence interval
    \begin{equation}
    \label{eq:pooled:CI}
      [Z_a + q_{\alpha/2}\hsigma_a,\; Z_a + q_{1-\alpha/2}\hsigma_a],
    \end{equation}
    where $q_p$ is the $p$-th quantile of the standard normal distribution.

Our second baseline is the \textbf{model-based metrics (MBM)} approach~\citep{miller2021modelbased}. As mentioned in \textsection\ref{sec:background}, MBM is a Bayesian approach to structured regression that models the scores produced by an AI system (like $\hY$ in the diabetes example). It is not directly applicable to performance metrics that are not based on scores, so we do not use it in the ASR experiments. Similar to SR, MBM uses linear modeling, and so requires specifying features for each data point. It comes with a boostrapping procedure for constructing confidence intervals.

We also compare our point estimates with the classical \textbf{James-Stein (JS)} estimator~\cite{james1961estimation,stein1956inadmissibility}.
The estimator works by shrinking standard estimates towards zero (or some other constant). We use a variant due to~\citet{bock1975minimax}, which is adapted to unequal variances (in our case, pooled estimates $\hsigma_a^2=\hsigma^2/n_a)$, giving rise to
    \[
      \hmu^\JS_a = \hmu_0 +
               \Parens{1-
                  \frac{(\card{\cA}-3)\hsigma^2}{\sum_{a'\in\cA}n_{a'}(Z_{a'}-\hmu_0)^2}
               }_+(Z_a-\hmu_0),
    \]
    where $\hmu_0=(\sum_{a\in\cA} n_a Z_a)/n$ is a weighted average of $Z_a$'s. Compared with Bock's original estimator~\citep{bock1975minimax}, we use $\card{\cA}$ in the numerator, as this has been previously observed to lead to better performance~\citep{feldman2012multi}.
Since $\hmu^\JS$ is not an unbiased estimator, construction of confidence intervals presents a challenge and we are not aware of any standard procedure.

\begin{figure*}%
\centering%
\includegraphics{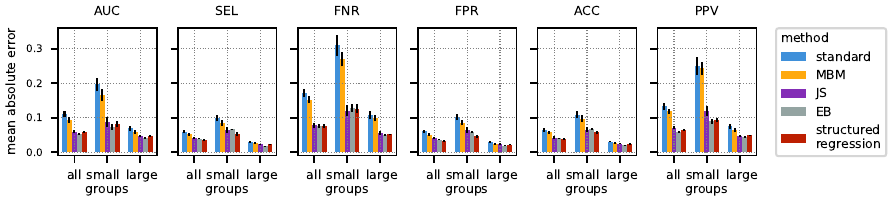}

\figsqueeze
\caption{\emph{Mean absolute error of estimates of 6 metrics using 5 methods on diabetes data.} Averaged across all groups, small groups (size at most 25), and large groups (size above 25), across 20 draws of evaluation dataset.}
\label{fig:diabetes-mae}
\end{figure*}

And finally, we compare our method with the \textbf{empirical Bayes (EB)} approach~\citep{efron1973stein}, which posits a hierarchical Bayesian model, and then estimates $\mu_a$ by posterior means, while fixing hyperparameters to their point estimates. In Appendix~\ref{appsec:EB} we derive the following variant, which we use in our experiments:
    \[
      \hmu^{\EB}_a = \hmu +
               \Parens{1-
                  \frac{\hsigma_a^2}{\htau^2+\hsigma_a^2}
               }(Z_a-\hmu),
    \]
    where $\hsigma^2_a$ is the pooled estimate of variance, and $\htau^2$ and $\hmu$ are obtained by
    \begin{align*}
      \htau^2 &= \Parens{\frac{\sum_{a\in\cA}n_{a}(Z_{a}-\hmu_0)^2 - (\card{\cA}-1)\hsigma^2}{n-\sum_a n_a^2/n}}_+,
    \\[6pt]
      \hmu &= \frac{\sum_{a\in\cA} Z_a/(\htau^2+\hsigma_a^2)}{\sum_{a\in\cA} 1/(\htau^2+\hsigma_a^2)}.
    \end{align*}
Similar to JS, we are not aware of any standard procedure for construction of confidence intervals.

\subsection{Diabetes experiments}
\label{sec:exp:diabetes}

We first explore the scenario from Example~\ref{ex:diabetes} using
the dataset developed by~\citet{strack2014impact}, and previously used in an AI fairness tutorial~\citep{gandhi2021fairness} and to evaluate the MBM approach~\citep{miller2021modelbased}. The dataset contains hospital admission records from 130 hospitals in the U.S.\ over a ten-year period (1998--2008) for patients who were admitted with a diabetes diagnosis and whose hospital stay lasted one to fourteen days.
It is a tabular dataset with 47 features describing each encounter, including patient demographics and clinical information.
Following \citet{miller2021modelbased}, we filter out records with missing demographics and those with age below~20. We preprocess clinical features as in~\citep{gandhi2021fairness}.
To emulate an AI system that scores patients for a post-discharge care program, we use 25\% of the data to train a logistic regression model to predict whether the patient will be readmitted into hospital within 30 days. The remaining 75\% of the data, consisting of 73,988 hospital admissions across 55,157 individuals, is used as the ground truth $\cD$ in all of our evaluation experiments.\looseness=-1

We consider three sensitive attributes, $\race$, $\age$, and $\gender$, with $\cA_\race=\set{\text{African American}, \text{Hispanic}, \text{white}, \text{other}}$,
$\cA_\age=\set{\text{20--39}, \text{40--59}, \text{60--79}, \text{80--99}}$, and $\cA_\gender=\set{\text{male}, \text{female}}$.
Hospital admissions are represented as tuples $(X,A,Y,\hY)$, where $X$ contains clinical features, $A=(\race,\age,\gender)$, $Y\in\set{0,1}$ indicates whether the patient was readmitted within 30 days of discharge, and $\hY\in[0,1]$ is the readmission probability predicted by the logistic regression model.
From the ground truth we then sample an evaluation dataset $S$ of size 5000 by stratified sampling according to $A$.\looseness=-1

As in Example~\ref{ex:diabetes}, we assume that the hospital uses a threshold~$r$, and patients with $\hY\ge r$ are automatically referred into the care program. We set the threshold $r$ so that $\Pr_{\cD}[\hY\ge r]=0.2$, meaning that only 20\% of patients are referred, and write $\pi(\hY)=\one\set{\hY\ge r}$ to denote this decision rule. We consider six performance metrics (including those already introduced earlier), defined for any $p\in\Delta$ as\looseness=-1
\begin{align*}
  \sel(p)&=\Pr_p\bracks{\pi(\hY)=1},
\\
  \acc(p)&=\Pr_p\bracks{\pi(\hY)=Y},
\\
  \fnr(p)&=\Pr_p\bracks{\pi(\hY)=0\given Y=1},
\\
  \fpr(p)&=\Pr_p\bracks{\pi(\hY)=1\given Y=0},
\\
  \ppv(p)&=\Pr_p\bracks{Y=1\given\pi(\hY)=1},
\\
  \auc(p)&=\Pr_{(Y,\hY)\sim p,\,(Y',\hY')\sim p}\bracks{\hY<\hY'\given Y=0,Y'=1}.
\end{align*}
The first five metrics (selection rate, accuracy, false positive rate, false negative rate, and positive predictive value) are derived from the confusion matrix. The final metric is the area under the ROC curve; $(Y,\hY)$ and $(Y',\hY')$ in its definition are sampled independently according to $p$.

In order to apply SR, we need to specify features $\vphia$. As sensitive features, we use indicators of \race, \age, \gender, as well as indicators of the triple (\race,\age,\gender). We use 7 explanatory features: indicators for 2 possible values of $Y$, and 5 additional clinical features describing the number of inpatient visits, outpatient visits, and emergency visits in the preceding year, number of diagnoses at admission, and whether any of the diagnoses was congestive heart failure. For MBM, we use the same set of features, but without the triple indicators.

In Figure~\ref{fig:errors-sel-fnr} from earlier, point estimates obtained by SR appear to be closer to the ground truth than those obtained by the standard method and MBM. Confidence intervals constructed by SR are of similar size as the standard confidence intervals, and occasionally smaller. MBM appears to produce smaller confidence intervals than SR, but they seem to miss the ground-truth metric values more often. We next evaluate these anecdotal observations more systematically.

\begin{figure*}%
\centering%
\includegraphics{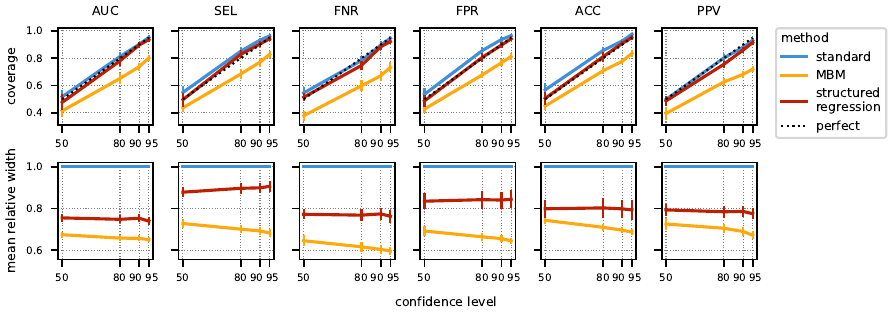}

\figsqueeze
\caption{\emph{Coverage and mean relative width of confidence intervals for 6 metrics constructed by 3 methods on diabetes data.} Averaged across all groups and across 20 draws of evaluation dataset. Relative width is with respect to the width of the standard confidence interval.\looseness=-1}
\label{fig:diabetes-cov}
\end{figure*}

In Figure~\ref{fig:diabetes-mae}, we evaluate the quality of point estimates using \emph{mean absolute error} (MAE), which is the mean deviation of the point estimate from the truth, averaged across 20 draws of evaluation dataset, and over all groups, or separately over the groups of size at most 25 (which we call \emph{small}) and groups of size greater than 25 (which we call \emph{large}).
JS, EB and SR yield substantially more accurate point estimates than the standard method and MBM. The improvement is particularly dramatic for small groups.
JS, EB and SR all perform at a similar level,
but SR tends to work best on small groups, and EB is marginally better than JS and SR on large groups (see Figure~\ref{fig:diabetes-mae-best} in Appendix~\ref{appsec:diabetes} for a comparison limited to these three methods). We use MAE instead of MSE, because MAE values are easier to interpret, but MSE results are qualitatively similar.\looseness=-1

In Figure~\ref{fig:diabetes-cov}, we shift attention to confidence intervals. In the top plots, we evaluate \emph{coverage}, that is, how often the ground truth lies in the confidence intervals (across 20 draws of evaluation dataset and across all groups). We show coverage as a function of the confidence level. We see that both standard method as well as SR are well-calibrated, with their coverage close to the confidence level, whereas MBM is over-confident, with coverage well below the confidence level. In the bottom plots, we evaluate the \emph{mean relative width} of confidence intervals, meaning the mean of the ratio between the width of a confidence interval and the width of the standard confidence interval.
We see that MBM has the narrowest intervals, but this is at the expense of coverage. On the other hand, SR is able to maintain well-calibrated coverage while still decreasing the confidence intervals by up to 20\% compared with the standard method.

\begin{table}
\caption{\emph{Goodness-of-fit tests on diabetes data.} From left to right, we consider increasingly more complex models with a growing set of features and report the $p$-values of the associated goodness-of-fit tests; $p$-values below 0.05 are in bold.\looseness=-1}
\label{tab:diabetes}
\tabsqueeze
\renewcommand{\tabcolsep}{4.5pt}
\centering
\small
\begin{tabular}[t]{lcccc}
\toprule
 Estimated & \multicolumn{4}{c}{Goodness-of-fit test $p$-values}
 \\
 metric & \multicolumn{4}{c}{(comparing a more expressive \textit{vs} a less expressive model)}
 \\
\cmidrule{2-5}
 & \multicolumn{1}{l}{\hphantom{\textit{vs} }$\expl$}
 & \multicolumn{1}{l}{\hphantom{\textit{vs} }$\sens$}
 & \multicolumn{1}{l}{\hphantom{\textit{vs} }$\expl + \sens$}
 & \multicolumn{1}{l}{\hphantom{\textit{vs} }$\expl + \sens + Y\! \cdot \sens$}
 \\
 & \multicolumn{1}{l}{\textit{vs }$\emptyset$}
 & \multicolumn{1}{l}{\textit{vs }$\emptyset$}
 & \multicolumn{1}{l}{\textit{vs }$\expl$}
 & \multicolumn{1}{l}{\textit{vs }$\expl + \sens$}
 \\
\midrule
AUC & \textit{0.281} & \textit{0.438} & \textit{0.726} & \textit{0.543} \\
SEL & \textbf{0.000} & \textbf{0.000} & \textbf{0.000} & \textit{0.842} \\
FNR & \textit{0.093} & \textit{0.473} & \textit{0.735} & \textit{0.565} \\
FPR & \textbf{0.001} & \textbf{0.000} & \textbf{0.000} & \textit{0.431} \\
ACC & \textbf{0.000} & \textbf{0.000} & \textbf{0.005} & \textit{0.182} \\
PPV & \textit{0.316} & \textit{0.493} & \textit{0.470} & \textit{0.874} \\
\bottomrule
\addlinespace
\multicolumn{5}{l}{\textit{Model abbreviations: $\emptyset$=intercept only, {\expl}=explanatory features,}}
\\
\multicolumn{5}{l}{\textit{\hphantom{Model abbreviations: }{\sens}=sensitive features, $\cdot$=interactions}}
\end{tabular}

\end{table}

Finally, in Table~\ref{tab:diabetes}, we demonstrate the use of goodness-of-fit tests.  We begin with the question: \textit{Is there statistically significant evidence of performance disparity across groups; and if so, is there further evidence of intersectional harm?} Table~\ref{tab:diabetes} shows $p$-values for goodness-of-fit tests beginning with just the intercept, adding explanatory features, then sensitive features (just the indicators of \race, \age, and \gender, but not of their combination), and eventually interaction terms between the outcome $Y$ and sensitive features. There is no evidence to go beyond the intercept-only model when estimating AUC, FNR, PPV. That is, there is no \textit{detectable} variation in AUC, FNR, or PPV across groups (or other explanatory variables).
For FNR, this is consistent with what we observe in Figure~\ref{fig:errors-sel-fnr}.  The confidence intervals shown are large and overlapping for the vast majority of groups, even after SR is applied to help reduce uncertainty.  Because of how wide the FNR confidence intervals are, the reasonable conclusion is that sample sizes are too small for the inference to be conclusive, and \textit{not} that we have definitive evidence of equal performance across groups.
On the other hand, the table shows that both explanatory and sensitive features help with modeling SEL, FPR, and ACC. In fact, sensitive features improve the fit \emph{after} the explanatory features have already been added, meaning that differences in performance across the groups cannot be explained by the ``benign'' explanatory features alone.

In Appendix~\ref{appsec:semi_synth_gof} we provide more examples of insight that may be gained through goodness-of-fit analysis using semi-synthetic variants of the diabetes data generated to exhibit different ground-truth structures for the underlying variation in system performance.

\begin{figure*}%
\centering%
\includegraphics{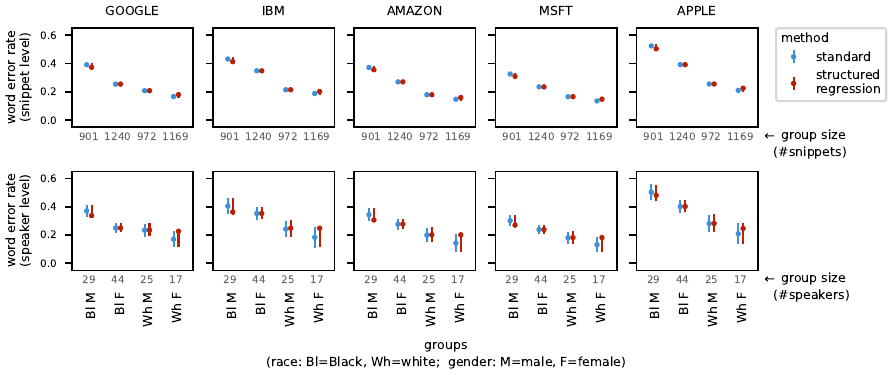}

\figsqueeze
\caption{\emph{Point estimates and 95\% confidence intervals of word error rates of five ASR systems.}}
\label{fig:errors-asr}
\end{figure*}

\subsection{Experiments with ASR data}
\label{sec:exp:ASR}

We now explore the scenario from Example~\ref{ex:ASR} using the data provided by \citet{koenecke2020racial} as a supplement to their paper finding racial disparities in commercial ASR systems. Similar to \citet{koenecke2020racial}, we use the \emph{matched} dataset, which contains 4282 snippets across 105 distinct speakers. (Matching ensures that there is the same number of snippets from Black and white speakers and that the marginal distributions of various descriptive statistics match.)

For each audio snippet, we are provided with various statistics (like duration and word count),
an anonymized speaker~id, speaker demographics, and word error rates (WERs) on that snippet by five ASR systems (by Google, IBM, Amazon, Microsoft, and Apple). This information is encoded as a tuple $(X,A,W_1,\dotsc,W_5)$, where $X$ contains the identity of the speaker, the duration of the snippet in seconds, and word count, $A$ contains two sensitive attributes, \gender and \race, $\cA_\gender=\set{\text{male},\text{female}}$ and $\cA_\race=\set{\text{Black}, \text{white}}$, and finally, instead of $Y$ (human transcription) and $\hY_1,\dotsc,\hY_5$ (transcriptions by five ASR systems), we directly have the corresponding word error rates $W_i=\weru(\hY_i,Y)$. The performance metric for the system $i$ is thus $m(p)=\E_p[W_i]$ for any $p\in\Delta$.

Although there appears to be a large number of samples ($n=4282$), there are only 105 distinct speakers.
We expect there to be a substantial amount of correlation between WERs of the same individual,
so an analysis that treats the WERs as independent is likely to overstate the statistical significance of findings, and may arrive at incorrect conclusions, in particular,
when some speakers have many more snippets than others.
In our experiments, we therefore present results both from a snippet-level analysis that treats the WERs across all snippets as independent (as done in \citep{koenecke2020racial}), and a speaker-level analysis that first reduces the data to speaker-level WERs by taking an average of WERs across the speaker's snippets.\looseness=-1

We first compare disaggregated evaluation results obtained by SR versus the standard method. To apply SR, we need to specify features $\vphia$. As sensitive features, we use indicators of \race and \gender, as well as indicators of the pair (\race,\gender). We use only one explanatory feature, equal to the log duration of the snippet.

In Figure~\ref{fig:errors-asr}, we report the results. At the snippet level, both methods generally replicate the results of \citet{koenecke2020racial}: Black male speakers have the largest WER, followed by Black female speakers, white male speakers, and white female speakers. The main difference is that SR systematically shrinks the WER values of the extreme groups (Black male speakers and white female speakers) towards the mean.
Results at the speaker level have substantially larger confidence intervals than the snippet-level results, reflecting smaller group sizes.
Also, due to smaller group sizes, the SR point estimates are shrunk towards the mean more aggressively.

We also carry out the goodness-of-fit analysis of structure of intersectional harms. At the speaker level, we find that the variation of performance of all systems is well-explained by the additive model $\expl+\race+\gender$ (the $p$-values of adding each variable in turn are below 0.003), but not by a model with interactions. This is in contrast with the snippet-level analysis, which supports the model with interactions (with $p$-values below 0.001). We interpret this conservatively and conclude that there is evidence for an additive structure of intersectional harms, but not for an interaction term. This does not mean that there are no interaction effects, just that we cannot conclude that from the data at hand.\looseness=-1

\section{Conclusion}
\label{sec:conclusion}

We have introduced a structured regression approach to disaggregated evaluation and compared its performance with a variety of baselines. We have seen that the structured regression (SR), James-Stein (JS) and empirical Bayes (EB) estimators all substantially improve accuracy of point estimates compared with the standard approach as well as a more sophisticated MBM baseline.
SR, JS and EB are simple to implement, and are also close in terms of performance, so the choice among them should be driven by their usability. Here, SR has some advantages. Its ability to include application-specific features makes it more flexible, and it has a well-developed inference procedures like construction of confidence intervals and goodness-of-fit tests.
Examining JS and EB more closely from inference perspective in the context of disaggregated evaluation is a promising direction for future research and a necessity for their practical use.
Note that we have evaluated SR only in two domains, so any applications in domains with different characteristics (like the number and types of explanatory and sensitive features, or dataset size) require additional validation.

Many important challenges lie outside the scope of this paper. For example, we assume that relevant sensitive attributes and performance metrics have been determined.
However, as \citet{barocas2021designing} discuss, the sensitive attributes often include socially constructed---and potentially contested---features (like race and gender), which makes the task of mapping people to attributes and corresponding subgroups potentially fraught, particularly when it involves inference or use of proxy variables, or poses a risk for members of already-marginalized subgroups.
As a separate challenge, in many high-stakes applications (like education and healthcare), we are not able to directly measure who might benefit, so we need to rely on proxies.
A poor choice of a proxy may further exacerbate existing inequities, as is the case, for instance, when predicting risk of re-offense from arrest records~\citep{fogliato2020fairness} or predicting healthcare needs based on healthcare expenditures~\citep{obermeyer2019dissecting}.\looseness=-1

Once the disaggregated results are produced, a complementary set of challenges arises in how to interpret them. We have conspicuously omitted analysis of regression coefficients, because in our preliminary experiments, we found that lasso coefficients exhibit too much variance for reliable inference. Instead, we suggest to use goodness-of-fit tests and have demonstrated several ways how. We acknowledge that we have just taken some initial steps in this area, and there are many opportunities to apply more sophisticated statistical techniques. Our exploration also completely leaves out important sociotechnical questions about how to draw actionable conclusions, and how to best communicate the results to relevant stake-holders, both of which are key in translating fairness assessments into a reduction in fairness-related harms.\looseness=-1

\begin{acks}
We would like to thank Misha Khodak for many helpful discussions and in particular for pointing out the connection between disaggregated evaluation and the James-Stein estimator.
\end{acks}

\begin{figure*}%
\centering%
\includegraphics{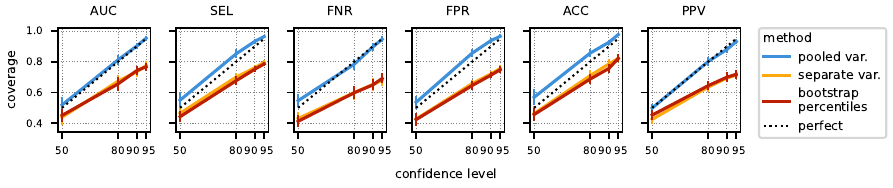}

\figsqueeze
\caption{\emph{Comparison of methods for constructing confidence intervals for the standard estimator.} Showing coverage of confidence intervals constructed for six metrics on diabetes data, averaged over all groups and over 20 draws of evaluation dataset. Confidence intervals constructed from pooled variance are close to the perfect line (corresponding to coverage equal to confidence level). Confidence intervals derived from separately estimated variances undercover true values.}
\label{fig:diabetes-naive-cov}
\end{figure*}

\bibliographystyle{ACM-Reference-Format}%
\bibliography{refs}

\appendix

\section{Confidence intervals for standard estimates}
\label{appsec:pooled}

We consider three methods for constructing confidence intervals for standard estimates $Z_a$ at a given confidence level~$\gamma$ (e.g., 95\%), or equivalently, at a significance level~$\alpha=1-\gamma$ (e.g., 5\%).

Two of the methods are based on normal approximation and take form
\[
   [Z_a + q_{\alpha/2}\hsigma_a,\; Z_a + q_{1-\alpha/2}\hsigma_a],
\]
where $q_p$ is the $p$-th quantile of the standard normal distribution and $\hsigma_a^2$ is an estimate of variance of $Z_a$. We consider either the pooled estimate of variance derived in \Eq{pooled}, or the estimate $(\hsigma_a^\boot)^2$ obtained by boostrap on $S_a$. The third method uses bootstrap percentiles on $S_a$.

In Figure~\ref{fig:diabetes-naive-cov}, we compare coverage properties of the resulting confidence intervals on diabetes data. Confidence intervals constructed from pooled variance estimates are well-calibrated, with coverage closely matching their confidence level. The other two methods substantially undercover true values.

\section{Derivation of the empirical Bayes estimator}
\label{appsec:EB}

We posit the following hierarchical Gaussian model:
\begin{equation}
\label{eq:hier:model}
\begin{aligned}
  \mu_a&\sim\cN(\mu,\tau^2)\quad\text{for all $a\in\cA$},
\\
  Z_a&\sim\cN(\mu_a,\sigma_a^2)\quad\text{for all $a\in\cA$},
\end{aligned}
\end{equation}
where $\sigma_a$ is known and $\mu$ and $\tau$ are unknown hyperparameters. We observe values $Z_a$ and need to predict $\mu_a$.

Conditioning on the prior and observations, we obtain the posterior distribution
\[
  \mu_a\given \mu,\tau,Z_a
  \sim\cN\Parens{\hmu^\EB_a, (\hsigma^\EB_a)^2}
\]
where the posterior mean and variance are equal to
\begin{equation}
\label{eq:EB:posterior}
  \hmu^\EB_a
  = \mu+\frac{\tau^2}{\tau^2+\sigma_a^2}\cdot(Z_a-\mu),
\quad
  (\hsigma^\EB_a)^2
  = \frac{\tau^2}{\tau^2+\sigma_a^2}\cdot\sigma_a^2.
\end{equation}

The empirical Bayes approach takes point estimates of the hyperparameters $\mu$ and $\tau$, and plugs them into \Eq{EB:posterior}. The resulting $\hmu^\EB_a$ is used as a point estimate for $\mu_a$ and the resulting $\hsigma^\EB_a$ is used to construct credible intervals for $\mu_a$.\looseness=-1

\begin{table*}
\caption{\emph{Semi-synthetic data generation.}
We consider four semi-synthetic ground-truth structures,
which we refer to as
$\model_\age$, $\model_{\age+\rc}$, $\model_{\age\cdot\rc}$, and $\model_\expl$, depending which variables they depend on and how, with ``$+$'' denoting additive dependence and ``$\cdot$'' presence of interaction.
Model coefficients have been chosen to ensure that in all cases $\E_\cD[\hY]\approx 0.27$ and $(\var_\cD[\hY])^{1/2}\approx0.44$.}
\label{tab:synth:generation}
\tabsqueeze
\small
\begin{center}
\begin{tabular}{lll}
\toprule
Model name
  & Ground-truth value of $\mu_a$
  & Data-generating process
\\
\midrule
$\text{model}_{\age}$
  & $\mu_a = 0.35 - 0.3\cdot\phia{\age,\text{40--60}}$
  & $\hY=\Bernoulli(\mu_A)$
\\
$\text{model}_{\expl}$
  & $\mu_a = -0.93
       + 0.16\cdot\E_{\cD_a}\bracks{X_{\textit{number\_diagnoses}}}$
  & $\hY=\cN(\mu_A,0.1)$
\\
$\text{model}_{\age+\rc}$
  & $\mu_a = 0.65
       - 0.15\cdot\phia{\age,\text{40--60}}
       - 0.45\cdot\phia{\race,\text{white}}$
  & $\hY=\Bernoulli(\mu_A)$
\\
$\text{model}_{\age\cdot\rc}$
  & $\mu_a = 0.32
       - 0.27\cdot\phia{\age,\text{40--60}}\cdot\phia{\race,\text{white}}$
  & $\hY=\Bernoulli(\mu_A)$
\\
\bottomrule
\end{tabular}
\end{center}
\end{table*}

\begin{table*}
\caption{\emph{Goodness-of-fit tests on semi-synthetic data.} From left to right, we consider increasingly more complex models with a growing set of features and report the $p$-values of the associated goodness-of-fit tests; $p$-values below 0.05 are in bold.}
\label{tab:synth}
\tabsqueeze
\small
\renewcommand{\tabcolsep}{4pt}
\begin{tabular}{lccccccc}
\toprule
 Data-generating & \multicolumn{7}{c}{Goodness-of-fit test $p$-values}
 \\
 model & \multicolumn{7}{c}{(comparing a more expressive \textit{vs} a less expressive model)}
 \\
\cmidrule{2-8}
 & \multicolumn{1}{l}{\hphantom{\textit{vs} }$\expl$}
 & \multicolumn{1}{l}{\hphantom{\textit{vs} }$\age$}
 & \multicolumn{1}{l}{\hphantom{\textit{vs} }$\expl{\,+\,}\age$}
 & \multicolumn{1}{l}{\hphantom{\textit{vs} }$\expl{\,+\,}\rc$}
 & \multicolumn{1}{l}{\hphantom{\textit{vs} }$\expl{\,+\,}\age{\,+\,}\rc$}
 & \multicolumn{1}{l}{\hphantom{\textit{vs} }$\expl{\,+\,}\age{\,+\,}\rc$}
 & \multicolumn{1}{l}{\hphantom{\textit{vs} }$\expl{\,+\,}\age{\,+\,}\rc{\,+\,}\age{\,\cdot\,}\rc$}
 \\
 & \multicolumn{1}{l}{\textit{vs }$\emptyset$}
 & \multicolumn{1}{l}{\textit{vs }$\emptyset$}
 & \multicolumn{1}{l}{\textit{vs }$\expl$}
 & \multicolumn{1}{l}{\textit{vs }$\expl$}
 & \multicolumn{1}{l}{\textit{vs }$\expl{\,+\,}\age$}
 & \multicolumn{1}{l}{\textit{vs }$\expl{\,+\,}\rc$}
 & \multicolumn{1}{l}{\textit{vs }$\expl{\,+\,}\age{\,+\,}\rc$}
 \\
\midrule
$\text{model}_{\age}$ & \textbf{0.025} & \textbf{0.000} & \textbf{0.000} & \textit{0.487} & \textit{0.153} & \textbf{0.000} & \textit{0.576} \\
$\text{model}_{\expl}$ & \textbf{0.000} & \textbf{0.000} & \textit{0.323} & \textit{0.366} & \textit{0.608} & \textit{0.551} & \textbf{0.000} \\
$\text{model}_{\age+\rc}$ & \textbf{0.013} & \textit{0.661} & \textit{0.142} & \textbf{0.000} & \textbf{0.000} & \textbf{0.000} & \textit{0.089} \\
$\text{model}_{\age\cdot\rc}$ & \textbf{0.003} & \textbf{0.000} & \textbf{0.000} & \textbf{0.040} & \textbf{0.015} & \textbf{0.000} & \textbf{0.002} \\
\bottomrule
\addlinespace
\multicolumn{8}{l}{\textit{Model abbreviations: $\emptyset$=intercept only, {\expl}=explanatory features, \rc=\textit{race}, $\cdot$=interactions}}
\end{tabular}

\end{table*}

We estimate $\tau^2$ by analyzing a suitable sum of squares (similarly as in the analysis of variance). To start, note that if we marginalize out $\mu_a$ from \Eq{hier:model}, we find that the values $Z_a$ are conditionally independent given $\mu$ and $\tau$, with
\begin{equation}
\label{eq:Za}
  Z_a\given\mu,\tau\sim\cN(\mu,\tau^2+\sigma_a^2)\quad\text{for all $a\in\cA$}.
\end{equation}
For each $a$, we consider the square $(Z_a-\hmu_0)^2$, where
\[
  \hmu_0=\sum_{a\in\cA}w_a Z_a,
\quad
  \text{with}
\quad
  w_a=n_a/n
  \text{ for all $a\in\cA$.}
\]
The expectation of $(Z_a-\hmu_0)^2$ then takes the following form:
\begin{align}
\notag
&\E\BigBracks{\BigParens{Z_a-\hmu_0}^2\BigGiven\mu,\tau}
\\
\notag
&\qquad{}
=
\E\BiggBracks{
     \BiggParens{\sum_{a'\ne a} w_{a'}Z_{a'}
     -(1-w_a)Z_a}^2 \BiggGiven\mu,\tau}
\\
\notag
&\qquad{}
=
\E\BiggBracks{
     \BiggParens{\sum_{a'\ne a} w_{a'}(Z_{a'}-\mu)
     -(1-w_a)(Z_a-\mu)}^2 \BiggGiven\mu,\tau}
\\
\label{eq:square}
&\qquad{}
=
\sum_{a'\ne a} w_{a'}^2(\tau^2+\sigma_{a'}^2) + (1-w_a)^2(\tau^2+\sigma_a^2)
\\
\label{eq:square:final}
&\qquad{}
=
\sum_{a'\in\cA} w_{a'}^2(\tau^2+\sigma_{a'}^2)
+(1-2w_a)(\tau^2+\sigma_a^2),
\end{align}
where \Eq{square} follows by \Eq{Za} and conditional independence of $Z_a$'s. Multiplying \Eq{square:final} by $w_a$ and summing across all $a$,
we obtain
\begin{align*}
&
\E\BiggBracks{
     \sum_{a\in\cA} w_a\BigParens{Z_a-\hmu_0}^2
     \BiggGiven\mu,\tau}
\\
&\qquad{}
=
  \sum_{a'\in\cA} w_{a'}^2(\tau^2+\sigma_{a'}^2)
     +\sum_{a\in\cA}w_a(1-2w_a)(\tau^2+\sigma_a^2)
\\
&\qquad{}
=
  \sum_{a\in\cA}w_a(1-w_a)(\tau^2+\sigma_a^2).
\end{align*}
We rearrange the final expression to obtain an unbiased estimate of $\tau^2$:
\begin{equation}
\label{eq:htau2}
  \htau^2 = \frac{\sum_{a\in\cA} w_a(Z_a-\hmu_0)^2 - \sum_{a\in\cA} w_a(1-w_a)\sigma_a^2}{1-\sum_{a\in\cA} w_a^2}.
\end{equation}
Since $\E[Z_a\given\mu,\tau]=\mu$ and $\var[Z_a\given\mu,\tau]=\tau^2+\sigma_a^2$, we estimate $\mu$ by taking a weighted average of $Z_a$'s, with the weights proportional to the inverse of the variance, but with $\htau^2$ plugged in for $\tau^2$:
\begin{equation}
\label{eq:hmu}
  \hmu = \frac{\sum_{a\in\cA} Z_a/(\htau^2+\sigma_a^2)}{\sum_{a\in\cA} 1/(\htau^2+\sigma_a^2)}.
\end{equation}
The last missing piece is $\sigma^2_a$, for which we use the pooled estimate from \Eq{pooled}.

Combining it all together, we use the pooled estimates of variance $\hsigma^2_a$ and the weighted mean $\hmu_0$ alongside observations $Z_a$ to calculate $\htau^2$ using \Eq{htau2}; then we calculate $\hmu$ using \Eq{hmu}; and finally we calculate $\hmu^\EB_a$ and $(\hsigma^\EB_a)^2$ using \Eq{EB:posterior}.

\section{Goodness-of-fit experiments with semi-synthetic data}
\label{appsec:semi_synth_gof}

In this appendix we provide further examples of the kinds of insight concerning the structure of intersectional harm that may be obtained through goodness-of-fit analysis.  Our examples rely on semi-synthetic data.  More precisely, we continue to use the diabetes dataset as described in \textsection\ref{sec:exp:diabetes}, but with different values $\hY$. We consider the performance metric $m(p)=\E_p[\hY]$ (this is quite similar to selection rate or word error rate) and generate $\hY$ in such a way that ground-truth metric values $\mu_a$ have a specific structure.

We consider four semi-synthetic ground-truth structures, which we refer to as $\model_\age$, $\model_{\age+\rc}$, $\model_{\age\cdot\rc}$, and $\model_\expl$, depending which variables they depend on and how, with ``$+$'' denoting additive dependence and ``$\cdot$'' presence of interactions.  These models correspond, respectively, to settings where the true variation in model performance is explained entirely by \textit{age} alone; by \textit{age} and \textit{race} additively, with middle-age patients and white patients experiencing lower values of the performance metric; by \textit{age} and \textit{race} interactionally, with the intersectional group of \textit{white middle-age} patients experiencing lower values of the performance metric whereas middle-age patients of other races and white patients of other ages do not; and entirely by \textit{non-demographic explanatory factors}.    The details of the underlying data generation process are summarized in Table~\ref{tab:synth:generation}.

Table~\ref{tab:synth} summarizes the results of a sequence of goodness-of-fit tests set up to investigate different questions about the structure of the underlying performance disparity using the observed semi-synthetic data.  Moving from left to right, we test goodness-of-fit of more and more complex models.  The first column, \textit{expl vs $\emptyset$} asks: \textit{is the data consistent with constant model performance that does vary with available (non-demographic) explanatory variables?} In the first row, the ground-truth depends only on \age, but we see a statistically significant improvement in goodness-of-fit from $\emptyset$ to $\expl$, because of correlation between $\age$ and $\expl$. The third column then asks: \textit{is the variation in performance explained by the benign explanatory factors alone, or is there evidence of further variation with age?} After the explanatory features have been included, $\age$ still helps (the improvement from $\expl$ to $\expl+\age$ is significant), so the variation in the performance metric cannot be explained by the ``benign explanatory factors'' alone.  While not presented here, the results of \textit{expl + age vs age} are \textit{not} statistically significant in this setting, confirming that the benign factors do not explain additional variation that is not explained by age alone.

Shown in the second row is the setting where the data is drawn from $\model_\expl$.  Columns 3 and 4 ask whether there is additional variation in performance in \textit{age} or \textit{race}, respectively, that is not explained by the benign factors.  The $p$-values for these goodness-of-fit tests are not statistically significant, so we conclude correctly that there is no evidence that $\age$ or $\race$ help after explanatory features have been added.

\begin{figure*}%
\centering%
\includegraphics{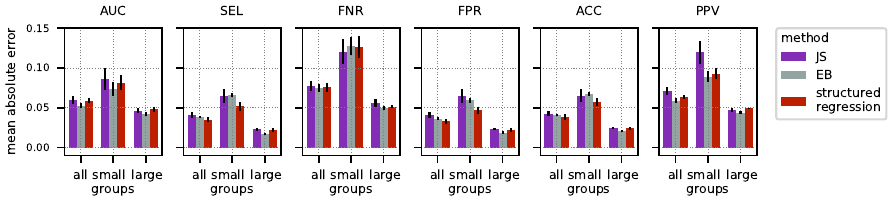}

\figsqueeze
\caption{\emph{Mean absolute error of estimates of 6 metrics using 3 best-performing methods on diabetes data.} Averaged across all groups, small groups (size at most 25), and large groups (size above 25), across 20 draws of evaluation dataset.}
\label{fig:diabetes-mae-best}
\end{figure*}

\begin{figure*}%
\centering%
\includegraphics{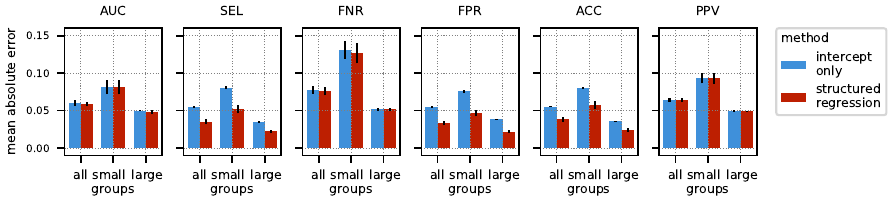}

\figsqueeze
\caption{\emph{Comparison of structured regression with the intercept-only model.}
Showing mean absolute error, averaged across all groups, small groups (size at most 25), and large groups (size above 25), across 20 draws of evaluation dataset.}
\label{fig:diabetes-intercept-only}
\end{figure*}

In the last two rows of Table~\ref{tab:synth}, we demonstrate how goodness-of-fit tests handle data from additive models versus models with interaction terms. The former correspond to the situation when harms experienced by intersectional groups combine additively, the latter when there is an additional intersectional effect. For the additive ground truth ($\model_{\age+\rc}$), tests suggest a sequence of variable additions $\expl+\rc+\age$, but then show no support for including interaction terms. For the data from $\model_{\age\cdot\rc}$, tests correctly provide support for an inclusion of interactions.  That is, we correctly identify the presence of super-additive harms that would accrue to intersectional age-race subgroups.

\section{Additional diabetes experiments}
\label{appsec:diabetes}

In Figure~\ref{fig:diabetes-mae-best}, we evaluate the quality of the point estimates produced by the three best-performing methods. In Figure~\ref{fig:diabetes-intercept-only}, we compare the performance of the structured regression approach with an intercept-only model.

\end{document}